\documentclass[11pt]{article}

\usepackage[letterpaper,margin=1in]{geometry}
\usepackage{natbib}
\usepackage{lmodern}

\usepackage[utf8]{inputenc}
\usepackage[T1]{fontenc}
\usepackage{hyperref}
\usepackage{url}
\usepackage{booktabs}
\usepackage{amsfonts}
\usepackage{amsmath,amssymb,amsthm}
\usepackage{nicefrac}
\usepackage{microtype}
\usepackage{xcolor}
\usepackage{bm}
\usepackage{bbold}
\usepackage{graphicx}
\usepackage[textsize=tiny]{todonotes}
\usepackage{enumitem}
\usepackage{authblk}
\usepackage{orcidlink}
\hypersetup{
    colorlinks=true,
    allcolors=blue!50!black,
}

\newcommand{\fourier}{\mathcal{F}}
\newcommand{\hilbert}{\mathcal{H}}

\newcommand{\sgn}{\operatorname{sgn}}

\newcommand{\propname}{CF-Prop}
\newcommand{\methodname}{CF-VI}

\title{Exact Bayesian layers via characteristic functions}

\author[1]{William H. W. Thompson\,\orcidlink{0000-0002-1893-240X}\thanks{\href{mailto:wthomps3@uvm.edu}{wthomps3@uvm.edu}}}
\author[2]{Guillaume St-Onge\,\orcidlink{0000-0001-5415-3748}}
\author[1,3,4]{Laurent Hébert-Dufresne\,\orcidlink{0000-0002-0008-3673}}
\author[1,3,5]{Jean-Gabriel Young\,\orcidlink{0000-0002-4464-2692}\thanks{\href{mailto:jean-gabriel.young@uvm.edu}{jean-gabriel.young@uvm.edu}}}
\affil[1]{Vermont Complex Systems Institute, University of Vermont}
\affil[2]{School of Electrical Engineering and Computer Science, University of Ottawa}
\affil[3]{Department of Computer Science, University of Vermont}
\affil[4]{Santa Fe Institute}
\affil[5]{Department of Mathematics and Statistics, University of Vermont}

\begin{document}

\maketitle

\begin{abstract}
    Bayesian neural networks provide principled uncertainty estimates, but training them with Monte Carlo sampling can be expensive.
    Sampling-free variational inference algorithms make training Bayesian neural networks feasible, but existing approaches assume a Gaussian output distribution, which can be inaccurate in applied problems.
    Here, we introduce \propname{}, an uncertainty propagation method that uses characteristic functions to compute the exact output distribution of a Bayesian ReLU layer, without any sampling, even for non-Gaussian predictive densities.
    Because \propname{} propagates characteristic functions, the evidence lower bound---the objective of variational training---can be computed in closed form for a broad class of likelihoods, including the Gaussian, Poisson, and Gamma.
    This allows us to implement an efficient variational inference algorithm, \methodname{}.
    On CIFAR-100, \methodname~achieves the lowest calibration error of any tested method, improving on the next best method by about 35\%.
    On a heavy-tailed insurance claim dataset, \methodname~accurately captures tail uncertainty that is severely underestimated by both other sampling-free methods and Monte Carlo methods.
\end{abstract}

\section{Introduction}

Neural networks are the least reliable when training data is scarce.
Bayesian Neural Networks (BNNs) offer a principled way to deal with the fundamental uncertainty of this data regime \citep{mitros_validity_nodate, kristiadi_being_2020}. 
They quantify both \emph{epistemic uncertainty}, which arises from limited knowledge of the model, and \emph{aleatoric uncertainty}, which is due to the inherent randomness of the data-generating process.
They excel in high-stakes, safety-critical applications \citep{ghoshal_estimating_2020, ding_capture_2021}, perform well with limited data, prevent overfitting \citep{depeweg_decomposition_2018}, and can increase fairness and reduce bias \citep{doherty_individual_2023}.

BNNs quantify epistemic uncertainty by introducing a posterior distribution $p(\boldsymbol{\theta} \mid \mathcal{D})$ over the model parameters $\boldsymbol{\theta}$, given training data $\mathcal{D}$.
Propagating this uncertainty to the output $\eta$ is required for training and inference. 
Stochastic methods, such as Bayes by Backprop \citep{graves_practical_nodate, blundell_weight_2015}, estimate it by sampling weights.
Sampling-free methods, such as Deterministic Variational Inference (DVI) \citep{wu_deterministic_2019}, CovProp \citep{wright_analytic_2024}, and Variational Bayesian Last Layers (VBLL) \citep{harrison_variational_2024}, replace sampling with moment propagation, which removes noise at the cost of assuming a specific functional form for the pre-activations.

We argue that these methods break down exactly where uncertainty quantification matters most.
Indeed, DVI, CovProp and VBLL assume that the predictive distribution and the variational family are Gaussians.
However, many applied problems require accurate estimates of tail probabilities---the probability of a 100-year flood or a multi-million-dollar insurance claim---for distributions that are often far from Gaussian.
This Gaussian assumption shows up in two places in these methods: the likelihood, which captures aleatoric uncertainty, and the distribution of the network, which captures epistemic uncertainty.
But in practice, the likelihood can be skewed, in Gamma regressions for claim sizes, for example.
And the distribution of the network output also tends to be skewed in these cases, because tail predictions are often dominated by a few rarely active neurons.
Hence, non-Gaussian variational families are required to properly handle these cases.
Any propagation method that models either distribution as Gaussian can severely underestimate tail risk.
In practice, the BNN literature uses many other variational families: spike-and-slab families for sparsification~\citep{Hubin2019}, scale mixtures for shrinkage~\citep{louizos_bayesian_2017, ghosh_model_2019}, and multimodal mixtures~\citep{dusenberry_efficient_2020, miller_variational_2017}---though current sampling-free propagation methods are incompatible with these families.

We introduce \propname{}, an exact and sampling-free method for uncertainty propagation in a single-layer Bayesian perceptron that uses characteristic functions (CF).
The forward pass of the layer is expressed in the Fourier domain, making the predictive density tractable. We use this propagation algorithm as the backbone of a new method, Variational Inference via Characteristic Functions (\methodname{}).
\methodname{} extends sampling-free variational inference beyond the Gaussian regime by expressing the predictive density of the Bayesian layer \emph{exactly} and in closed form. It allows for both non-Gaussian predictive distributions and non-Gaussian variational families. 
It can accommodate any likelihood and any variational family with a tractable characteristic function.

\methodname{} can be used to train a fully exact Bayesian perceptron or as an exact Bayesian last layer on a sampling-free or stochastic trunk.
We show that \methodname{} performs well on traditional benchmarks such as UCI regressions, and provides state-of-the-art uncertainty quantification on classification tasks with a large number of labels such as CIFAR-100.
We also show that \methodname{} performs exceptionally well in uncertainty quantification for neural GLMs commonly used in the actuarial sciences.
In particular, we show that for a Gamma regression task on heavy-tailed data---predicting the size of insurance claims---other methods severely underestimate uncertainty in the tails of the distribution.
In contrast, our method provides an exact description of the tail uncertainty.
\methodname{} is also computationally efficient: In many common cases, it adds only a modest overhead to the training of a fully deterministic neural network. 

Our \textbf{contributions} are as follows:
\begin{enumerate}[topsep=0pt, itemsep=1pt, parsep=0pt]
    \item An exact expression for the output distribution of a Bayesian ReLU layer using characteristic functions.
    \item \methodname{}, a sampling-free variational inference method compatible with any mean-field variational family and a broad class of likelihoods.
    \item Lower training and inference cost than other sampling-free methods on UCI benchmarks, at comparable NLL.
    \item The best calibration among tested methods on CIFAR-100.
    \item Accurate tail uncertainty on heavy-tailed regression, where other tested methods underweight tail risk.
\end{enumerate}

\section{Background: Uncertainty propagation}
\label{sec:background}

Given a distribution $q(\boldsymbol{\theta})$ over the full set of model parameters $\boldsymbol{\theta}$, uncertainty propagation computes the induced distribution of the network output at an input $x_n$, $q(\eta \mid x_n)$, the \emph{latent predictive distribution}, which is needed both to train a BNN and to make predictions. The latent predictive distribution quantifies epistemic uncertainty in the prediction due to uncertainty in the model parameters.
The model also includes a likelihood $p(y_n \mid \eta)$, which links the latent output $\eta$ to the observation $y_n$ and models the \emph{aleatoric uncertainty}, the irreducible noise in the data.
Combining the two gives the \emph{posterior predictive distribution}, $q(y_n \mid x_n) = \mathbb{E}_{\eta \sim q(\eta \mid x_n)}[p(y_n \mid \eta)] = \int p(y_n \mid \eta)\, q(\eta \mid x_n)\, d\eta$.

Directly calculating the probability density function of the activations of a BNN is so computationally expensive that it is rarely implemented.
Instead, training and inference of BNNs rely on stochastic sampling \citep{graves_practical_nodate, blundell_weight_2015} or sampling-free approximations.
We review sampling-free methods below. 

\subsection{Moment propagation} 

Even when the full density $q(\eta \mid x_n)$ is intractable, its \emph{moments} might be easy to compute. Moment propagation methods derive exact or approximate expressions for the moments of the activations of a single hidden layer, then moment-match, assuming that the distribution of the activations is well described by a Gaussian sharing these moments.

Two methods are relevant here: deterministic variational inference (DVI) and covariance propagation (CovProp).
\citet{wu_deterministic_2019} develop DVI expressions for the first and second moments of the activation distribution and derive an asymptotic approximation for the correlation between activations.
DVI generally tracks the full covariance structure among all $h$ hidden units, with $\mathcal{O}(h^3)$ complexity.
A diagonal method (dDVI) can also be implemented by ignoring covariance, reducing DVI's cubic complexity with layer width to linear complexity. 

\citet{wright_analytic_2024} extend this work.
CovProp expresses the moments as a power series in the correlation coefficient.
The resulting method can, in principle, attain an arbitrarily accurate representation of the first two moments, provided that enough terms are computed in the power series.
In practice, CovProp requires at least 4 terms to outperform DVI. 

Both DVI and CovProp can be applied to arbitrarily deep networks, although approximation error compounds for deeper networks: even after 5 layers, DVI's predictive distribution is distorted significantly.
Competitive benchmark performance is attained only for very shallow networks of 1 or 2 layers \citep{wu_deterministic_2019}. 

\subsection{Variational Bayesian last layers} 

Variational Bayesian last layers (VBLLs)~\citep{harrison_variational_2024} are another sampling-free approach.
This is an example of Bayesian last-layer methods~\citep{kristiadi_being_2020}, which have many of the benefits of a fully Bayesian neural network.
They use a deep deterministic ``trunk'' to generate low-dimensional features, which are then fed as inputs to a shallow Bayesian network.
VBLLs are a sampling-free variational approach where only the linear readout is Bayesian.

While VBLLs are efficient, they have limited expressiveness. VBLLs capture uncertainty only in the linear readout; they cannot account for uncertainty through a full non-linear layer.
Like moment propagation methods, VBLLs restrict the predictive distribution of the output $\eta$ to a Gaussian, and thus cannot account for any skew or heavy tails.
A VBLL classifier, for example, can only express uncertainty around decision boundaries that are linear in the learned features.
While highly efficient, the linear method cannot properly forecast uncertainty when data is sparse.

\subsection{Failure of Gaussian assumptions}

Moment propagation is a good approximation only when two assumptions hold: the likelihood is Gaussian, and the latent predictive distribution $q(\eta \mid x_n)$ is Gaussian.
Many common scenarios violate one or both, including typical configurations of the UCI regression benchmarks frequently used to evaluate Bayesian neural networks.

First, we consider the assumption that the likelihood is Gaussian.
If it is not, the expected log-likelihood $\mathbb{E}_{\eta \sim q(\eta \mid x_n)}[\log p(y_n \mid \eta)]$ is no longer determined by the first two moments of $q(\eta \mid x_n)$.
For the Poisson and Gamma likelihoods with a log link, the expectation contains exponential moments, $\mathbb{E}[e^{\eta}]$ and $\mathbb{E}[e^{-\eta}]$.
These are sensitive to the tails and only weakly constrained by the first two moments.

The second assumption, that the latent predictive is Gaussian, rests on the central limit theorem.
When the linear readout sums many weakly correlated activations together, the predictive will be approximately Gaussian.
\citet{wu_deterministic_2019} argue that the correlation between hidden units remains weak enough during training for this to hold approximately, and in a heteroscedastic regression with two hidden layers of 128 units they show close agreement between the Monte Carlo sampled and the moment-matched Gaussian activations.
However, many commonly used architectures have last layers with fewer than 128 units: the network used later in the same paper for the UCI regression benchmark has a single hidden layer of 50 units.

\begin{figure}[t]
  \centering
  \includegraphics[width=\textwidth]{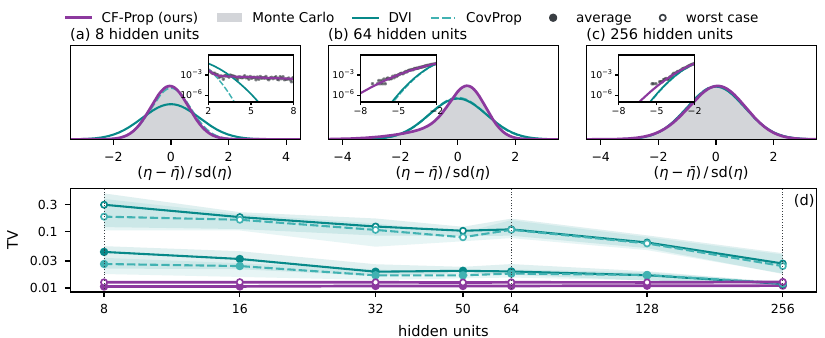}
  \caption{\textbf{Predictive tails are non-Gaussian for moderate-width networks.} A network with two hidden layers of $h$ units each (a deterministic layer followed by a Bayesian layer) is trained on UCI Energy with a 90/10 held-out split, using \methodname{} with a mean-field Gaussian variational family and prior and a Gaussian likelihood. 20 networks are trained per width.
  \textbf{Top}: z-scored predictive density of \propname{} (purple), DVI (teal, solid), CovProp (teal, dashed), and Monte Carlo (grey) with layers of \textbf{(a)}  8, \textbf{(b)} 64, and \textbf{(c)} 256 units.
  The insets compare the tails up to $8\sigma$ from the mean.
  \textbf{Bottom}: Total variation (TV) distance between the sampling-free approximation and Monte Carlo predictions, showing \textbf{(d)} the mean (filled) and worst-case (maximum, open) TV over test points.}
  \label{fig:act_skewed}
\end{figure}

Layers of this size (50 units) show meaningfully non-Gaussian predictive distributions, especially in the tails.
In Figure~\ref{fig:act_skewed}, we train networks with a deterministic hidden layer followed by a Bayesian hidden layer using \methodname{} with a mean-field Gaussian family, varying the layer's width from 8 to 256 units, and apply \propname{}, DVI, and CovProp to the same trained network to isolate the differences that can be attributed to propagation alone.
All methods model the mean correctly, but DVI and CovProp underestimate the tails.
Averaged over test inputs, their error is small, but at individual inputs it can be large.
At the width of the example in \citet{wu_deterministic_2019} (128 units), the average total variation between DVI and Monte Carlo is 0.017, while its worst case over inputs is 0.064.
\propname{} matches the Monte Carlo baseline throughout. 

The worst-case failures occur where training data is scarce.
Figure~\ref{fig:epistemic_gap} below shows that moment propagation works well in distribution but fails to generalize correctly.
Moment propagation is thus liable to fail in low-data regimes, where epistemic uncertainty quantification is most important.

\section{\propname{}: Exact uncertainty propagation via characteristic functions}
\label{sec:propagation}

We develop an exact description of the predictive density for a perceptron with ReLU activations.
We express the forward pass of a network in the Fourier domain via characteristic functions, the Fourier transforms of probability density functions.
This approach builds on prior work from \citet{Pilipovsky2023}, which expresses the action of the non-linear activation function in the Fourier domain using Hilbert transforms.
That work was focused on the output density of a deterministic perceptron with random input.
Here, we adapt the method to calculate the predictive density of a random network with a deterministic input.
Our network consists of a hidden layer of ReLU units with random weights and biases, followed by a deterministic linear readout.
It can be used on its own, as a trained perceptron, or as a Bayesian last layer. 

\subsection{Characteristic functions}

\emph{Characteristic functions} can be used to propagate the full distribution of uncertainty.
The CF $\phi_{X}(t)$ of a random variable $X$ is defined as:
\begin{equation}
    \phi_{X}(t) = \mathbb{E}[e^{\mathrm{i}tX}] = \int_{\mathbb{R}} e^{\mathrm{i}tx} f_X(x) \, dx
\end{equation}
where $t$ is a real number, and $f_X(x)$ is the probability density function of $X$. 
It uniquely characterizes the random variable $X$.

The CF of a sum of independent random variables is the product of their CFs.
Combined with the scaling rule $\phi_{aX}(t) = \phi_X(at)$ for a constant $a$, the CF of a linear combination $S_n = \sum_{i=1}^n a_i X_i$ of independent random variables is thus the product
\begin{equation}
    \phi_{S_n}(t) = \prod_{i=1}^n \phi_{X_i}(a_i t) \; .
    \label{eq:cf_sum}
\end{equation}

\subsection{Bayesian ReLU layer}
We consider a single hidden layer with a deterministic linear readout,
\begin{equation}
    \boldsymbol{\eta} = \mathbf{V}\,\mathrm{ReLU}(\mathbf{z}) + \mathbf{c}, \qquad \mathbf{z} = \mathbf{W}\mathbf{x} + \mathbf{b},
\end{equation}
where $\mathbf{x}$ is the deterministic $d \times 1$ input to the layer, $\mathbf{W}$ is an $h \times d$ weight matrix, $\mathbf{b}$ is an $h \times 1$ bias vector, and $\mathbf{z}$ is the vector of \emph{pre-activations}.
The readout weights $\mathbf{V}$, an $o \times h$ matrix, and bias $\mathbf{c}$, an $o \times 1$ vector, are deterministic, where $o$ is the dimension of the output.

We model the rows $(\mathbf{W}_{u\cdot}, b_u)$ as mutually independent under $q$, meaning that the pre-activations $z_u$ are independent given $\mathbf{x}$.
When the entries within each row are also independent, as in a mean-field family, Eq.~\eqref{eq:cf_sum} gives the CF of the pre-activation of the $u$\textsuperscript{th} neuron:
\begin{equation}
    \phi_{z_u}(t) = \phi_{b_u}(t) \prod_{j=1}^{d} \phi_{W_{uj}}(x_j t) \; .
\end{equation}

\subsection{Feeding the CF through ReLU activations}

To go from pre-activations to the final readout, we must apply $\mathrm{ReLU}(z_u) = \max(0, z_u)$ and feed the results through the linear readout.
The non-linearity introduced by the ReLU poses a challenge when working with CFs, but it can be handled using \emph{Hilbert transforms} of the CF, as shown by \citet{Pilipovsky2023}.

More precisely, we find that the CF of the transformed pre-activation $\phi_{\text{ReLU}(z_u)}(t)$ is:
\begin{equation}
    \phi_{\mathrm{ReLU}(z_u)}(t)
    = \frac{1}{2}\Big\{ 1 + \phi_{z_u}(t)
    + \mathrm{i}\big( \hilbert[\phi_{z_u}](t) - \hilbert[\phi_{z_u}](0) \big) \Big\}
    \label{eq:relu_cf}
\end{equation}
where $\hilbert[\cdot](t)$ denotes the Hilbert transform, i.e., the convolution of $\phi(t)$ with the function $\frac{1}{\pi t}$, known as a Cauchy kernel.
See Appendix~\ref{app:relu} for details. 

Since $z_u$ depends only on the weights and bias of unit $u$, which are independent across units, the activations are independent given $\mathbf{x}$.
Hence, we can apply Eq.~\eqref{eq:cf_sum} again, with the readout weights $V_{ku}$ as the coefficients, to obtain the CFs of the outputs:
\begin{equation}
    \phi_{\eta_k}(t) = e^{\mathrm{i}tc_k} \prod_{u=1}^{h} \phi_{\mathrm{ReLU}(z_u)}(V_{ku}\, t),
    \label{eq:preceptron_cf}
\end{equation}
where $e^{\mathrm{i}tc_k}$ is the CF of the deterministic bias $c_k$.

Computations in this CF representation involve up to three steps: Evaluating the CFs of the weights, applying the Hilbert transforms in Eq.~\eqref{eq:relu_cf}, and computing an inverse Fourier transform of $\phi_{\eta_k}(t)$ to obtain the \emph{density} of the output,
\begin{equation}
    q(\eta_k \mid \mathbf{x}) = \frac{1}{2\pi} \int_{\mathbb{R}} e^{-\mathrm{i}t\eta_k}\, \phi_{\eta_k}(t) \, dt \; .
    \label{eq:cf_inversion}
\end{equation}
How many of these steps can be done in closed form depends on the choice of variational family.
For some choices of weight distributions like the Gaussian, the first two steps are analytic (see Appendix~\ref{app:relu} and the next section). For a general family, all three must be done numerically.
Importantly, the layer can be trained on $\phi_{\eta_k}$ directly, with no Fourier inversion, for the common likelihoods treated in Sec.~\ref{sec:cfvi}.
The inversion in Eq.~\eqref{eq:cf_inversion} is only required when the full density is needed for a calculation: for example, the selection criteria for points often depend on the full posterior predictive distribution rather than the first two moments. 

\begin{figure}[!t]
  \centering
  \includegraphics[width=\textwidth]{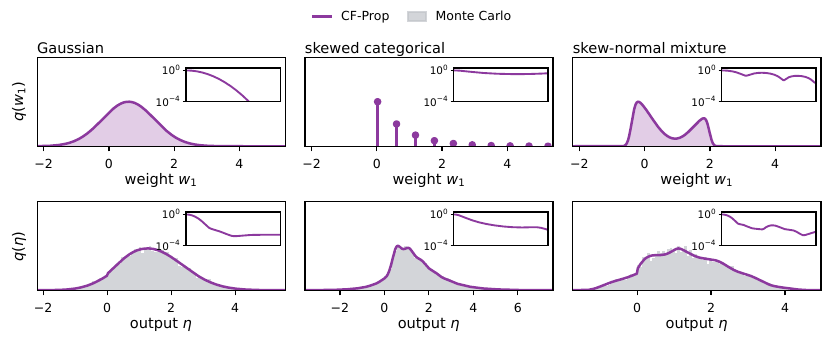}
  \caption{
    \textbf{Uncertainty propagation with \propname{} across variational families.}
    \textbf{Top}: Marginal density of a single weight $w_1$ for a Gaussian (left), skewed categorical (middle) and skew-normal mixture (right) family.
    The insets show the modulus of the CF of the weight, $|\phi_{w_1}(t)|$. 
    \textbf{Bottom}: The resulting latent predictive density $q(\eta)$ calculated with \propname{} (purple) compared to Monte Carlo samples (grey, $S=10^4$).
    The insets show the modulus of the CF of the output, $|\phi_\eta(t)|$.
    In all three cases, the CFs and densities are computed numerically with Fast Fourier transforms and their inverse.
  }
  \label{fig:families_toy}
\end{figure}

Figure~\ref{fig:families_toy} showcases the flexibility of \propname{}.
We propagate uncertainty from three different variational families: a Gaussian, a discrete skewed categorical, and a mixture of skew-normals.
In all three cases, \propname{} produces an estimate of the predictive density in close agreement with the Monte Carlo estimates.

\subsection{Closed-form weight distribution}
\label{sec:varfam}

Equation~\eqref{eq:preceptron_cf} has a closed-form expression in certain cases. 
For instance, when the weights are Gaussian, the pre-activations are linear combinations of independent Gaussians, and thus follow a Gaussian distribution themselves.
More precisely, if $W_{uj} \sim \mathcal{N}(\mu_{uj}, \sigma_{uj}^2)$ and $b_u \sim \mathcal{N}(\mu_{b_u}, \sigma_{b_u}^2)$, we have $z_u \sim \mathcal{N}(m_u, s_u^2)$ with
\begin{equation*}
    m_u = \sum_j \mu_{uj} x_j + \mu_{b_u}, \qquad
    s_u^2 = \sum_j \sigma_{uj}^2 x_j^2 + \sigma_{b_u}^2,
\end{equation*}
and $\phi_{z_u}(t) = e^{\mathrm{i} m_u t - \frac{1}{2} s_u^2 t^2}$.
In this case, one can show from Eq.~\eqref{eq:relu_cf} that the CF of the activation is
\begin{equation}
    \phi_{\mathrm{ReLU}(z_u)}(t) = \frac{1}{2}\operatorname{erfc}\!\Bigg(\frac{m_u}{\sqrt{2}\,s_u}\Bigg) + \frac{1}{2}\,\phi_{z_u}(t)\,\operatorname{erfc}\!\Bigg(-\frac{m_u + \mathrm{i}s_u^2 t}{\sqrt{2}\,s_u}\Bigg),
    \label{eq:relu_cf_gaussian}
\end{equation}
where $\operatorname{erfc}(\cdot)$ is the complementary error function, $\operatorname{erfc}(z) = 1 - \operatorname{erf}(z) = \frac{2}{\sqrt{\pi}} \int_z^\infty e^{-\xi^2}d\xi$, extended to complex arguments (see Appendix~\ref{app:relu} for details).

\section{\methodname{}: Variational Inference via Characteristic Functions}
\label{sec:cfvi}

\begin{figure}[h]
  \centering
  \includegraphics[width=0.9\textwidth]{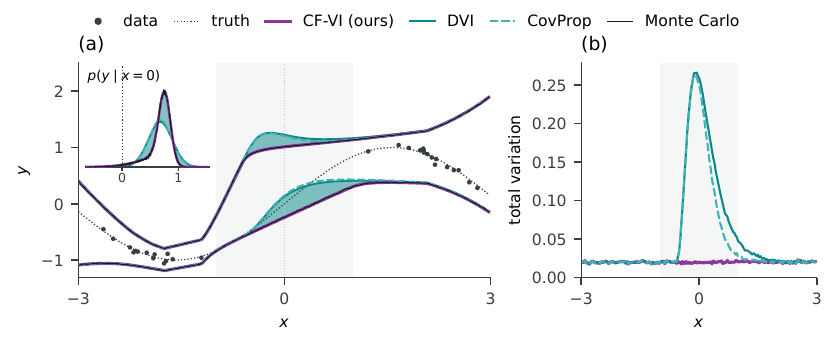}
  \caption{
    \textbf{\methodname\ improves epistemic uncertainty quantification in non-Gaussian regression problems.} 
    Synthetic data are generated as $y=\sin(x) + \epsilon$, where $\epsilon \sim \mathcal{N}(0, 0.05^2)$ is random.
    Training pairs are censored in the $x\in[-1,1]$ region, and a network (deterministic ReLU layer of 32 units, Bayesian ReLU layer of 32 units, linear readout, heteroscedastic Gaussian likelihood) is trained on 30 pairs with \methodname{}.
    \textbf{(a)} \methodname, DVI, CovProp, and Monte Carlo sampling ($10^5$ samples) are used to draw from the posterior predictive distribution of the same trained network. 
    The discrepancy between the 99\% credible interval and the Monte Carlo-sampled posterior, the reference, is shown as a shaded region for each method.
    The Monte Carlo reference is under the purple line.
    Inset: comparison of predictive distributions at $p(y | x = 0)$.
    \textbf{(b)} Total variation distance between the sampling-free and the Monte Carlo predictive distributions, as a function of $x$.
    Where training data is available, all three methods closely agree with the Monte Carlo estimate of the predictive density.
    In the censored region (grey band), however, only \methodname{} agrees with the MC samples, while DVI and CovProp underestimate uncertainty.
  }
  \label{fig:epistemic_gap}
\end{figure}

In Sec.~\ref{sec:propagation}, we outlined a general method for uncertainty propagation with characteristic functions.
We now use the method to build a variational inference algorithm for the Bayesian ReLU layer: \methodname{}.
We show how the exact expression for the output distribution yields an evidence lower bound (ELBO) that requires no sampling and is available in closed form for common likelihoods.

\methodname{} improves epistemic uncertainty quantification in non-Gaussian settings.
Figure~\ref{fig:epistemic_gap} compares the predictions of \methodname{}, DVI and CovProp for one network, trained with \methodname{}, on a toy regression problem with a gap in the training inputs.
Where training data is available, all three agree with the Monte Carlo reference.
Out of sample, however, DVI and CovProp underestimate epistemic uncertainty, and only \methodname{} remains in agreement with the reference.

\subsection{Variational objective}

Given a dataset $\mathcal{D} = \{(x_n, y_n)\}_{n=1}^{N}$, a likelihood $p(y_n \mid x_n, \boldsymbol{\theta})$ and a prior $p(\boldsymbol{\theta})$, variational inference approximates the posterior $p(\boldsymbol{\theta} \mid \mathcal{D})$ with a distribution $q(\boldsymbol{\theta})$ from a family $\mathcal{F}$ by maximizing the ELBO \citep{blei_variational_2017}:
\begin{equation}
  \operatorname*{arg\,max}_{q \in \mathcal{F}} \;
  \underbrace{\mathbb{E}_{q(\boldsymbol{\theta})}\!\left[ \log p(\mathcal{D} \mid \boldsymbol{\theta}) \right]}_{\text{reconstruction cost}}
  - \underbrace{D_{\mathrm{KL}}\!\left( q(\boldsymbol{\theta}) \,\|\, p(\boldsymbol{\theta}) \right)}_{\text{complexity cost}} \; .
\end{equation}
The complexity cost can be calculated analytically for suitable choices of prior and variational family.
For example, with the mean-field Gaussian family of Sec.~\ref{sec:varfam} and an independent $\mathcal{N}(0, \sigma_p^2)$ prior on each parameter, it is
\begin{equation}
  D_{\mathrm{KL}}\!\left( q(\boldsymbol{\theta}) \,\|\, p(\boldsymbol{\theta}) \right)
  = \frac{1}{2} \sum_{w} \left( \frac{\sigma_{w}^2}{\sigma_p^2} + \frac{\mu_{w}^2}{\sigma_p^2}
  - 1 - \log \frac{\sigma_{w}^2}{\sigma_p^2} \right) \; ,
  \label{eq:example_complexity_cost}
\end{equation}
where the sum is over all random weights $W_{uj}$ and biases $b_u$, and $\mu_w$ and $\sigma_w^2$ are the variational mean and variance of parameter $w$.
For other mean-field families, the complexity cost is cheap to compute numerically.

The reconstruction cost is a high-dimensional integral over the weights.
Because the likelihood depends on $\boldsymbol{\theta}$ only through the network output $\eta$, it can be rewritten as an integral over the output:
\begin{equation}
  \mathbb{E}_{q}\!\left[ \log p(\mathcal{D} \mid \boldsymbol{\theta}) \right]
  = \sum_{n} \int \log p(y_n \mid \eta)\, q(\eta \mid x_n)\, d\eta \; ,
  \label{eq:likelihood_cost_integral}
\end{equation}
where $q(\eta \mid x_n)$ is the distribution of the output induced by $q(\boldsymbol{\theta})$.
Section~\ref{sec:propagation} gives the CF of this distribution; we obtain $q(\eta \mid x_n)$ from the CF using an inverse Fast Fourier Transform (IFFT), and evaluate the integral in Eq.~\eqref{eq:likelihood_cost_integral} by quadrature.
Both operations are differentiable, so the ELBO can be optimized by backpropagation.
That said, for common likelihoods, the integral reduces to evaluations of the CF at a few points with no inversion, as we show next.

\subsection{Closed-form reconstruction cost}

For many likelihoods, we can circumvent the IFFT and quadrature by evaluating the reconstruction cost directly with the CF.
This is because the CF can be used to obtain two kinds of expectations directly.
Its derivatives at zero are the moments, $\mathbb{E}[\eta^{r}] = \mathrm{i}^{-r} \phi^{(r)}_{\eta \mid x_n}(0)$, and its values on the imaginary axis give exponential moments, $\mathbb{E}[e^{a\eta}] = \phi_{\eta \mid x_n}(-\mathrm{i}a)$.
Both are special cases of the relationship
\begin{equation}
    \mathbb{E}_{q(\eta \mid x_n)}\!\left[\eta^{r} e^{a\eta}\right] = \mathrm{i}^{-r}\, \phi^{(r)}_{\eta \mid x_n}(-\mathrm{i}a) \; ,
    \label{eq:cf_expectation}
\end{equation}
which follows from differentiating the definition of the CF $r$ times.
We call a likelihood an \emph{exponential-polynomial} if its logarithm is a sum of such terms,
\begin{equation}
    \log p(y \mid \eta) = \sum_{j=1}^{J} \alpha_j(y)\, \eta^{r_j} e^{a_j \eta} \; ,
    \label{eq:exp_sum_ll}
\end{equation}
where $r_j$ is a non-negative integer, $a_j$ can be a complex number, and the coefficients $\alpha_j(y)$ do not depend on $\eta$ but may depend on $y$.
Applying Eq.~\eqref{eq:cf_expectation} to the terms gives
\begin{equation*}
    \mathbb{E}_{q(\eta \mid x_n)}[\log p(y_n \mid \eta)] = \sum_{j=1}^{J} \alpha_j(y_n)\, \mathrm{i}^{-r_j}\, \phi^{(r_j)}_{\eta \mid x_n}(-\mathrm{i}a_j) \; ,
\end{equation*}
so the ELBO can be computed with at most $J$ evaluations of Eq.~\eqref{eq:preceptron_cf} and its derivatives per data point, whenever Eq.~\eqref{eq:exp_sum_ll} holds.

These likelihoods are exponential families whose natural parameters and log-partition function are exponential polynomials in $\eta$.
This exponential-polynomial family includes the Gaussian as a special case and, with a log link, the Poisson, the Gamma, the wider Tweedie family, and the Weibull.
The categorical distribution commonly used in classification is, unfortunately, not part of this family, but following \citet{harrison_variational_2024} we can derive a closed-form lower bound on the reconstruction cost, which we can optimize instead.
These results are derived in Appendix~\ref {app:ell}; see Table~\ref{tab:ell} for a summary.

Together with Eq.~\eqref{eq:example_complexity_cost}, our calculations yield an ELBO that can be computed in closed form for every exponential-polynomial likelihood, using only a small number of evaluations of the predictive CF and its derivatives.

\begin{table}[htbp]
\centering
\small
\caption{
    \textbf{Reconstruction costs in terms of characteristic functions}. We denote $\phi_\eta \equiv \phi_{\eta \mid x_n}$ for simplicity.
    The Gamma likelihood is expressed with a log link with shape $\nu$, and the Weibull uses a shape parameter $\kappa$.\\
}
\label{tab:ell}
\begin{tabular}{ll}
\toprule
Likelihood & $\mathbb{E}_q[\log p(y \mid \eta)]$ \\
\midrule
Gaussian & $-\tfrac{1}{2}\log(2\pi\sigma^2) - \big[y^2 + 2\mathrm{i}\,y\,\phi_\eta'(0) - \phi_\eta''(0)\big] / (2\sigma^2)$ \\
Poisson & $-\mathrm{i}\,y\,\phi_\eta'(0) - \phi_\eta(-\mathrm{i}) - \log y!$ \\
Gamma & $\nu \log \nu - \log \Gamma(\nu) + (\nu-1) \log y + \mathrm{i}\,\nu\,\phi_\eta'(0) - \nu\,y\,\phi_\eta(\mathrm{i})$ \\
Weibull & $\log\kappa + (\kappa-1)\log y + \mathrm{i}\,\kappa\,\phi_\eta'(0) - y^{\kappa}\,\phi_\eta(\mathrm{i}\kappa)$ \\
\midrule
Categorical & $\ge -\mathrm{i}\,\phi_{\eta_y}'(0) - \log \sum_{k'} \phi_{\eta_{k'}}(-\mathrm{i})$ \\
Any density & $\approx \Delta \eta \sum_{j} \log p(y \mid \eta_j)\, \fourier^{-1}[\phi_\eta](\eta_j)$ \\
\bottomrule
\end{tabular}
\end{table}

\section{Results}
\label{sec:experiments}

We evaluate \methodname{} in three settings: a Gamma neural GLM for insurance claim severity (Sec.~\ref{sec:exp_glm}), the UCI regression benchmarks (Sec.~\ref{sec:exp_uci}), and classification with Bayesian heads on frozen CIFAR-10 and CIFAR-100 features (Sec.~\ref{sec:exp_clf}).
On standard benchmarks, \methodname{} matches or exceeds the best sampling-free methods at lower prediction and training cost.
On the neural GLM task, it is the only tested method that does not underestimate the tail. 

\subsection{Case study: neural GLMs}
\label{sec:exp_glm}

Many applied problems use non-Gaussian likelihoods: Poisson and Gamma likelihoods for neural GLMs in actuarial science and forecasting \citep{noll_case_2018, wuthrich_statistical_2023, holvoet_neural_2025}.
Weibull likelihoods are used in neural survival analysis \citep{martinsson_wtte_2017, nagpal_deep_2021}.
All of these settings share heavy-tailed distributions where Gaussian approximations severely underestimate the risk of rare but catastrophic events. 

\begin{figure}[!b]
  \centering
  \includegraphics[width=\textwidth]{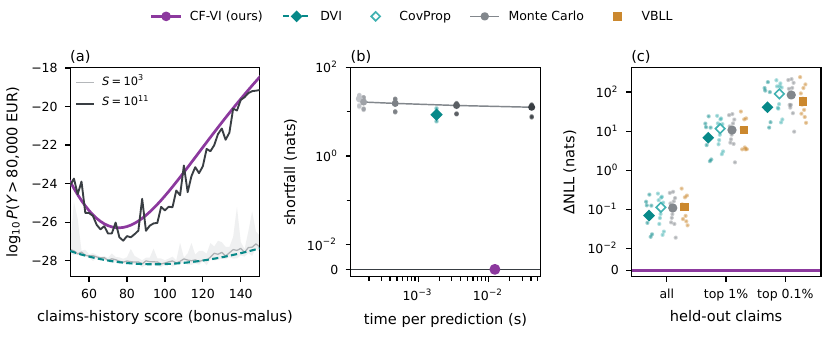}
  \caption{
    \textbf{\methodname{} outperforms sampling-free methods for heavy-tailed uncertainty quantification.}
    These results use the Gamma neural GLM of Sec.~\ref{sec:exp_glm}.
    \textbf{(a)} Log$_{10}$ probability of a claim above 80,000 EUR as a function of the bonus-malus score.
    Monte Carlo estimates are calculated with $10^3$ and $10^{11}$ samples (averaged across 8 runs in the $S=10^3$ case).
    They converge to the \methodname{} estimate slowly, and agree with DVI only when very few samples are used.
    \textbf{(b)} Held-out log-likelihood shortfall relative to \methodname{} on the top 1\% of claims, against prediction time (Monte Carlo: $S=10$ to $10^5$).
    \textbf{(c)} Held-out NLL minus that of \methodname{} (median over 10 seeds).
    Monte Carlo uses $S=100$ samples. 
  }
  \label{fig:severity}
\end{figure}

We consider an example taken from the freMTPL2 claim severity dataset of French motor insurance \citep{noll_case_2018}.
In this task, we predict the claim amount from policy features such as vehicle age and power, driver age, and location.
The distribution of claim sizes is very heavy-tailed.
As is typical for this setting, we use a Gamma likelihood with a network comprising one deterministic ReLU layer of 16 units, a Bayesian ReLU layer with 16 units and a mean-field Gaussian variational family, and a deterministic linear readout.
We report scores on held-out test sets using 5-fold cross-validation. 
Where applicable, we train 10 random seeds per fold.
The results are shown in Fig.~\ref{fig:severity}.

DVI severely underestimates the weight of the tail, and Monte Carlo estimates take a long time to converge.
Fig.~\ref{fig:severity}a shows the log probability that the fitted model assigns to large claims, over 80,000 EUR.
Holding all other features constant, we sweep the claim-history score (bonus-malus): a score below 100 indicates a driver with few previous claims, while a score above 100 indicates an accident-prone driver.
For high-risk individuals (a score of 140), both DVI and Monte Carlo sampling with $S=10^3$ underestimate the probability of a claim above 80,000 EUR by a factor of $10^8$.
Only by dramatically increasing the number of samples to $S=10^{11}$ does the Monte Carlo estimate converge to the \methodname{} estimate. 

In Fig.~\ref{fig:severity}b we evaluate the shortfall---the average difference in NLL between a method and \methodname{} on the top 1\% of claims by size---using Monte Carlo with $10$ to $10^5$ samples.
Despite the nearly $280\times$ increase in time, the increased number of samples reduces the shortfall by only 36\%, and at every sample size, MC performs worse than DVI.

\methodname{} also outperforms Monte Carlo ($S=100$) and moment-propagation methods in terms of predictive performance.
Fig.~\ref{fig:severity}c shows the difference in held-out negative log-likelihood (NLL) between \methodname{} and the other methods.
\methodname{} beats all competing methods by a median of 0.069 (DVI) to 0.115 (VBLL) nats per claim.
Its lead is strengthened on the largest claims: on the top 0.1\% of claims by size, \methodname{} beats the other methods by 40.7 (DVI) to 89.6 (CovProp) nats per claim.

These results suggest \methodname{} is valuable whenever forecasting tail risk is important, as in insurance claims, disaster forecasting, or survival analysis.
\methodname{} extends the use of Bayesian neural networks to these regimes. 

\subsection{Case study: UCI regression}
\label{sec:exp_uci}

\begin{figure}[h]
  \centering
  \includegraphics[width=\textwidth]{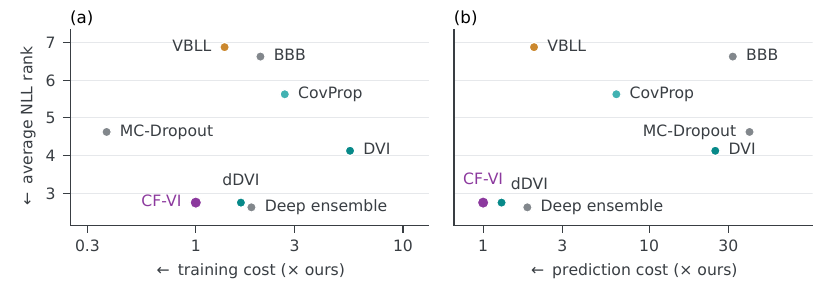}
  \caption{
    \textbf{Predictive performance versus cost on UCI regression.}
    Average rank by test NLL across datasets (lower is better) against the cost \textbf{(a)} per training step and 
    \textbf{(b)} per prediction of the test split.
    Costs are relative to \methodname{} and measured on one CPU core.
  }
  \label{fig:uci_pareto}
\end{figure}

Next, we compare our method to competing approaches on the UCI regression benchmarks.
The sampling-free methods most comparable to ours are DVI, dDVI \citep{wu_deterministic_2019}, CovProp \citep{wright_analytic_2024}, and VBLL \citep{harrison_variational_2024}.
We also include popular stochastic methods: Bayes by Backprop (BBB)~\citep{blundell_weight_2015}, deep ensembles~\citep{lakshminarayanan_simple_nodate}, and MC-Dropout~\citep{gal_dropout_2016, folgoc_is_2021}. 
We use the same training and test methods as \citet{wu_deterministic_2019} for DVI and as \citet{wright_analytic_2024} for CovProp. 

\begin{table}[t]
\centering
\caption{\textbf{UCI Regression benchmark.} We compute both the negative log-likelihood and the root mean square error on the test set, with standard errors.
Methods above the bar are sampling-free, methods below are stochastic. 
The best overall method and the best sampling-free methods are highlighted for each dataset.\\
}
\small
\setlength{\tabcolsep}{4pt}
\resizebox{\textwidth}{!}{%
\begin{tabular}{lccccccccc}
\toprule
Test NLL ($\downarrow$) & \textsc{Boston} & \textsc{Concrete} & \textsc{Energy} & \textsc{Kin8nm} & \textsc{Naval} & \textsc{Power} & \textsc{Wine} & \textsc{Yacht} & Avg.\ rank \\
\midrule
\methodname & $\mathbf{2.47 \pm 0.06}$ & $\mathbf{3.02 \pm 0.03}$ & $\mathbf{0.62 \pm 0.13}$ & $-1.284 \pm 0.007$ & $-6.15 \pm 0.03$ & $\mathbf{2.77 \pm 0.01}$ & $0.97 \pm 0.03$ & $\mathbf{0.25 \pm 0.05}$ & $2.75$ \\
DVI & $2.56 \pm 0.04$ & $3.10 \pm 0.02$ & $1.23 \pm 0.10$ & $-1.282 \pm 0.006$ & $-6.72 \pm 0.04$ & $2.79 \pm 0.01$ & $0.94 \pm 0.01$ & $0.52 \pm 0.06$ & $4.13$ \\
dDVI & $2.55 \pm 0.04$ & $3.08 \pm 0.02$ & $0.99 \pm 0.06$ & $\mathbf{-1.286 \pm 0.006}$ & $\mathbf{-6.85 \pm 0.03}$ & $2.78 \pm 0.01$ & $\mathbf{0.93 \pm 0.01}$ & $0.34 \pm 0.06$ & $2.75$ \\
CovProp & $2.72 \pm 0.03$ & $3.10 \pm 0.02$ & $1.57 \pm 0.07$ & $-1.246 \pm 0.007$ & $-3.94 \pm 0.25$ & $2.81 \pm 0.01$ & $0.95 \pm 0.01$ & $0.62 \pm 0.05$ & $5.63$ \\
VBLL & $3.21 \pm 0.17$ & $3.16 \pm 0.04$ & $0.70 \pm 0.03$ & $0.679 \pm 0.616$ & $-3.00 \pm 3.11$ & $4.17 \pm 0.45$ & $1.00 \pm 0.02$ & $1.58 \pm 0.28$ & $6.88$ \\
\midrule
Deep ensemble & $2.68 \pm 0.12$ & $3.03 \pm 0.10$ & $\mathbf{0.38 \pm 0.07}$ & $\mathbf{-1.337 \pm 0.008}$ & $\mathbf{-6.96 \pm 0.02}$ & $\mathbf{2.69 \pm 0.02}$ & $1.58 \pm 0.46$ & $\mathbf{-0.04 \pm 0.06}$ & $2.63$ \\
MC-Dropout & $2.59 \pm 0.07$ & $\mathbf{2.88 \pm 0.04}$ & $1.22 \pm 0.01$ & $-1.164 \pm 0.004$ & $-5.33 \pm 0.01$ & $2.78 \pm 0.01$ & $0.96 \pm 0.01$ & $1.62 \pm 0.02$ & $4.63$ \\
BBB & $2.72 \pm 0.02$ & $3.23 \pm 0.01$ & $1.54 \pm 0.10$ & $-1.242 \pm 0.006$ & $-4.55 \pm 0.05$ & $2.82 \pm 0.01$ & $0.96 \pm 0.01$ & $2.36 \pm 0.05$ & $6.63$ \\
\bottomrule
\end{tabular}}
\par\vspace{8pt}
\resizebox{\textwidth}{!}{%
\begin{tabular}{lccccccccc}
\toprule
Test RMSE ($\downarrow$) & \textsc{Boston} & \textsc{Concrete} & \textsc{Energy} & \textsc{Kin8nm} $\times10^{2}$ & \textsc{Naval} $\times10^{3}$ & \textsc{Power} & \textsc{Wine} & \textsc{Yacht} & Avg.\ rank \\
\midrule
\methodname & $\mathbf{3.28 \pm 0.18}$ & $\mathbf{5.23 \pm 0.11}$ & $\mathbf{0.47 \pm 0.01}$ & $\mathbf{7.10 \pm 0.06}$ & $2.15 \pm 0.17$ & $\mathbf{3.95 \pm 0.05}$ & $\mathbf{0.63 \pm 0.01}$ & $0.89 \pm 0.08$ & $2.75$ \\
DVI & $3.70 \pm 0.17$ & $5.89 \pm 0.13$ & $1.97 \pm 0.20$ & $7.19 \pm 0.05$ & $2.13 \pm 0.11$ & $4.00 \pm 0.05$ & $0.64 \pm 0.01$ & $1.59 \pm 0.32$ & $6.25$ \\
dDVI & $3.61 \pm 0.16$ & $5.82 \pm 0.14$ & $1.33 \pm 0.17$ & $7.14 \pm 0.05$ & $1.73 \pm 0.11$ & $3.98 \pm 0.05$ & $0.64 \pm 0.01$ & $1.05 \pm 0.09$ & $4.38$ \\
CovProp & $3.91 \pm 0.16$ & $5.86 \pm 0.12$ & $2.10 \pm 0.16$ & $7.47 \pm 0.06$ & $6.91 \pm 0.32$ & $4.05 \pm 0.05$ & $0.64 \pm 0.01$ & $1.24 \pm 0.10$ & $7.13$ \\
VBLL & $3.63 \pm 0.24$ & $5.32 \pm 0.13$ & $0.48 \pm 0.01$ & $10.90 \pm 1.23$ & $\mathbf{0.51 \pm 0.04}$ & $5.02 \pm 0.66$ & $0.64 \pm 0.01$ & $\mathbf{0.73 \pm 0.07}$ & $4.63$ \\
\midrule
Deep ensemble & $3.30 \pm 0.17$ & $4.94 \pm 0.12$ & $\mathbf{0.39 \pm 0.01}$ & $\mathbf{6.77 \pm 0.05}$ & $2.68 \pm 0.05$ & $\mathbf{3.78 \pm 0.05}$ & $0.64 \pm 0.01$ & $0.93 \pm 0.09$ & $2.88$ \\
MC-Dropout & $\mathbf{3.04 \pm 0.14}$ & $\mathbf{4.40 \pm 0.13}$ & $0.52 \pm 0.02$ & $7.30 \pm 0.04$ & $0.77 \pm 0.01$ & $3.87 \pm 0.05$ & $\mathbf{0.63 \pm 0.01}$ & $1.24 \pm 0.10$ & $2.75$ \\
BBB & $3.44 \pm 0.16$ & $5.90 \pm 0.12$ & $1.13 \pm 0.13$ & $7.05 \pm 0.05$ & $2.65 \pm 0.11$ & $4.04 \pm 0.05$ & $\mathbf{0.63 \pm 0.01}$ & $2.16 \pm 0.11$ & $5.25$ \\
\bottomrule
\end{tabular}}
\label{tab:uci_main}
\end{table}
 
We evaluate methods over 20 random 90/10 train/test splits for all UCI benchmarks.
All networks consist of two hidden layers of 50 units on a heteroscedastic Gaussian likelihood that outputs both the prediction $y(x)$ and a log-variance $\log \sigma^2(x)$. 
Architectures are as follows:
\begin{itemize}[topsep=0pt, itemsep=1pt, parsep=0pt]
    \item Full-covariance DVI, diagonal-covariance dDVI, and CovProp: two fully Bayesian layers and a deterministic readout, following the methods in the original papers.
    \item VBLL: two deterministic hidden layers with a VBLL readout.
    \item \methodname: deterministic hidden layer followed by a Bayesian hidden layer for the prediction $y(x)$. A separate deterministic hidden layer is used for the log-variance. 
\end{itemize}
Every variational method is trained with a mean-field Gaussian family with a standard normal prior.
BBB and MC-Dropout predictions use 100 samples, and the deep ensemble has five members.
Complete results are shown in Table~\ref{tab:uci_main}. 

Figure~\ref{fig:uci_pareto} shows the average predictive performance rank against cost per training step and per prediction for these methods. 
We find that \methodname{} is among the most efficient, costing only about 1.8 times as much per training step as the same network without the Bayesian layer.
Only MC-Dropout, which produces a much cruder uncertainty estimate, is cheaper but costs 40 times as much per prediction.
In terms of performance, \methodname{} ties with dDVI (avg rank 2.75) and is beaten by the deep ensemble (average rank 2.63), which costs 1.9 times as much per training step and per prediction.
We note, however, that while \methodname{} beats dDVI in cost, both algorithms have the same asymptotic complexity, and the efficiency difference likely reflects implementation details.
That said, we note that we provide an exact estimate of the full posterior predictive distribution as efficiently as dDVI provides its first two moments. 

\subsection{Case study: Classification}
\label{sec:exp_clf}

\begin{table}[htbp]
\centering
\caption{\textbf{CIFAR classification benchmark.} We show the mean expected calibration error (ECE), computed with 15 bins of equal width, the negative log-likelihood, and the accuracy for each method on CIFAR-100 and CIFAR-10, with standard deviations.
Methods above the bar are sampling-free, methods below are stochastic. The best overall method and the best sampling-free methods are highlighted for each dataset.\\
}

\small
\setlength{\tabcolsep}{4pt}
\resizebox{\textwidth}{!}{%
\begin{tabular}{lcccccc}
\toprule
 & \multicolumn{3}{c}{CIFAR-100} & \multicolumn{3}{c}{CIFAR-10} \\
\cmidrule(lr){2-4}\cmidrule(lr){5-7}
Head & ECE (\%) $\downarrow$ & NLL $\downarrow$ & Acc.\ (\%) $\uparrow$ & ECE (\%) $\downarrow$ & NLL $\downarrow$ & Acc.\ (\%) $\uparrow$ \\
\midrule
\methodname & $\mathbf{3.14 \pm 0.01}$ & $0.790 \pm 0.004$ & $81.27 \pm 0.16$ & $2.45 \pm 0.11$ & $0.173 \pm 0.006$ & $96.12 \pm 0.09$ \\
DNN & $5.56 \pm 0.30$ & $\mathbf{0.782 \pm 0.008}$ & $\mathbf{81.70 \pm 0.25}$ & $2.50 \pm 0.09$ & $0.169 \pm 0.005$ & $\mathbf{96.16 \pm 0.08}$ \\
VBLL & $5.43 \pm 0.73$ & $0.818 \pm 0.007$ & $81.58 \pm 0.23$ & $1.88 \pm 0.09$ & $0.154 \pm 0.005$ & $96.12 \pm 0.09$ \\
 Last-Layer Laplace & $29.29 \pm 0.64$ & $1.109 \pm 0.008$ & $81.69 \pm 0.28$ & $\mathbf{1.32 \pm 0.11}$ & $\mathbf{0.149 \pm 0.004}$ & $\mathbf{96.16 \pm 0.08}$ \\
DVI  & $7.94 \pm 0.71$ & $0.847 \pm 0.011$ & $81.34 \pm 0.14$ & $1.45 \pm 0.11$ & $0.159 \pm 0.005$ & $96.13 \pm 0.09$ \\
CovProp & $7.97 \pm 0.71$ & $0.847 \pm 0.011$ & $81.35 \pm 0.14$ & $1.44 \pm 0.12$ & $0.159 \pm 0.005$ & $96.13 \pm 0.09$ \\
\midrule
Deep ensemble & $8.42 \pm 0.38$ & $0.880 \pm 0.011$ & $81.56 \pm 0.31$ & $3.17 \pm 0.12$ & $0.249 \pm 0.008$ & $96.06 \pm 0.10$ \\
MC-Dropout & $7.43 \pm 0.28$ & $1.070 \pm 0.018$ & $81.35 \pm 0.26$ & $3.03 \pm 0.14$ & $0.293 \pm 0.006$ & $96.08 \pm 0.15$ \\
BBB & $4.84 \pm 0.55$ & $0.809 \pm 0.008$ & $81.23 \pm 0.18$ & $1.90 \pm 0.10$ & $0.160 \pm 0.006$ & $96.06 \pm 0.11$ \\
\bottomrule
\end{tabular}}
\label{tab:cifar}
\end{table}

We next evaluate \methodname{} on image classification, as a Bayesian head on a pretrained network.
We use the same set of baselines, with some additions and modifications; we include a strong last-layer Laplace method~\citep{daxberger_laplace_2021}, and adapt DVI and CovProp to work with a classification objective.
Unlike in regression, where the output is deterministic, in classification deep neural networks also output a categorical distribution over classes, so they are meaningful comparators and included here. 
Finally, we use the discriminative version of the VBLL classifier introduced in \cite{harrison_variational_2024}. 

For our trunk, we freeze a Wide ResNet-28-10 backbone \citep{zagoruyko_wide_2016} and train the methods as heads on its 640-dimensional features.
The \methodname{} head has one Bayesian hidden layer of 64 ReLU units, with a Gaussian variational family that has a full covariance over the incoming weights of each unit, and a $\mathcal{N}(0, 1/640)$ prior on all weights.
It is trained with the lower bound on the reconstruction cost given in Table~\ref{tab:ell}. 
For the DNN, we use the same architecture but with a deterministic hidden layer.

Table~\ref{tab:cifar} reports the results.
On CIFAR-100, \methodname{} has the lowest expected calibration error (ECE) of all methods: 3.1\%, against 4.8\% for the next best, Bayes by Backprop, a reduction of 35\%.
Its NLL and accuracy are also competitive.
On CIFAR-10, all heads are well calibrated, and \methodname{} does not perform as well.
According to these experiments, the benefits of \methodname{} are the most apparent in many-class problems.

\section{Conclusion}\label{sec:when}
We have used characteristic functions to derive an exact representation of the full predictive distribution of a Bayesian ReLU layer with a deterministic input and readout, and exact variational objectives for non-Gaussian likelihoods.
As demonstrated on CIFAR heads on a Wide ResNet backbone and on freMTPL2, \methodname{} excels as a Bayesian last layer on an ordinary deterministic network.  

In many cases, our method both captures uncertainty more accurately and can be run more efficiently than existing methods for moment propagation, providing a new optimal balance of predictive performance and efficiency in traditional classification and regression benchmarks.
It also extends sampling-free variational inference to non-Gaussian settings.
On freMTPL2 insurance severity, \methodname{} provided a far superior measure of tail uncertainty for rare but catastrophic predictions.
In doing so, it unlocks the application of sampling-free Bayesian neural networks to many applied settings where uncertainty quantification is critical.

Our approach also extends sampling-free mean-field VI to many other variational families and likelihoods, making deterministic variational inference compatible with a much wider array of architectures and methods. 

The greatest limitation of the current approach is its inability to capture correlations between activations.
Future work may extend these results using low-rank covariance structure to efficiently and exactly track correlations between layers.
This extension would be crucial in order to characterize the epistemic, out-of-distribution uncertainty, which mean-field variational families struggle to capture \citep{foong_-between_2019}.
Other integral transforms like the Mellin transform, in combination with similar low-rank correlation structures, may be able to describe the full forward pass of the network through many layers. 

The present work provides a foundation and proof of concept for integral-transform and Fourier-domain expressions of Bayesian neural networks, which can make BNNs useful across a much wider range of fields.  

\section{Code Availability}

The code for \propname{} and \methodname{} is available under an MIT license at \url{https://github.com/WillHWThompson/cf-vi}.

\bibliographystyle{plainnat}

\appendix

\section{Derivation of the ReLU characteristic function}
\label{app:relu}

This appendix derives Eq.~\eqref{eq:relu_cf} and its closed form for Gaussian pre-activations, Eq.~\eqref{eq:relu_cf_gaussian}.
Throughout, $X$ is a random variable with density $f_X(x)$ and CF $\phi_X(t) = \int_{\mathbb{R}} e^{\mathrm{i}tx} f_X(x)\, dx$.

\subsection{Characteristic function of the rectified variable}

We start by splitting the density of $X$ into the positive and negative parts of its support,
\begin{equation*}
    f_+(x) = f_X(x)\, \Theta(x), \qquad f_-(x) = f_X(x)\, \Theta(-x),
\end{equation*}
where $\Theta$ is the Heaviside step function, which can be written as $\Theta(\pm x) = \tfrac{1}{2}[1 \pm \sgn(x)]$, giving
\begin{equation}
    f_\pm(x) = \frac{1}{2} \big[ f_X(x) \pm \sgn(x)\, f_X(x) \big].
    \label{eq:f_pm}
\end{equation}
The first term of Eq.~\eqref{eq:f_pm} straightforwardly has the transform
\begin{equation*}
    \frac{1}{2}\int_{\mathbb{R}} e^{\mathrm{i}tx} f_X(x)\, dx = \frac{1}{2}\phi_X(t),
\end{equation*}
so most of the difficulty comes from the second term.

We use the Hilbert transform $\hilbert[\cdot]$ to compute this second term.
When applied to the CF, the Hilbert transform multiplies the density by $-\mathrm{i}\sgn(x)$,
\begin{equation*}
    \hilbert[\phi_X](t) = \int_{\mathbb{R}} e^{\mathrm{i}tx} \big[ -\mathrm{i}\sgn(x) \big] f_X(x)\, dx.
\end{equation*}
Hence, we can see that the Fourier transform of $\sgn(x) f_X(x)$ is  $\mathrm{i}\,\hilbert[\phi_X](t)$.
This lets us define the following quantities for the two parts of  Eq.~\eqref{eq:f_pm}:
\begin{equation*}
    \phi_\pm(t) := \int_{\mathbb{R}} e^{\mathrm{i}tx} f_\pm(x)\, dx = \frac{1}{2}\bigl[ \phi_X(t) \pm \mathrm{i}\, \hilbert[\phi_X](t) \bigr].
\end{equation*}

With these results in place, we can then compute the CF of the rectified variable $\mathrm{ReLU}(X)$.
The ReLU activation maps the negative part of the support of $X$ to a point mass at zero, with weight $P(X \le 0)$.
Noting that
\begin{equation*}
     P(X \le 0) = \int_\mathbb{R} f_-(x)\,dx = \phi_-(0),
\end{equation*}
since $e^{\mathrm{i}tx} = 1$ at $t = 0$, the CF of the rectified variable is given by
\begin{align*}
    \phi_{\mathrm{ReLU}(X)}(t) 
    &= \int_{\mathbb{R}} e^{\mathrm{i}tx} \big[f_+(x)+ \delta(x) \phi_{-}(0)\big]\, dx\\
    &= \phi_+(t) + \phi_-(0)\\
    &= \frac{1}{2}\big[\phi_X(t) + \mathrm{i}\, \hilbert[\phi_X](t)\big] + \frac{1}{2}\big[\phi_X(0) - \mathrm{i}\, \hilbert[\phi_X](0)\big].
\end{align*}
Finally, since $\phi_X(0) = 1$, we obtain Eq.~\eqref{eq:relu_cf} of the main text:
\begin{equation}
    \phi_{\mathrm{ReLU}(X)}(t) = \frac{1}{2}\Big[ 1 + \phi_X(t) + \mathrm{i}\big( \hilbert[\phi_X](t) - \hilbert[\phi_X](0) \big) \Big].
    \label{eq:relu_cf_app}
\end{equation}

\subsection{Gaussian pre-activations}

When the pre-activation is Gaussian, the Hilbert transforms in Eq.~\eqref{eq:relu_cf_app} can be calculated in closed form.
The CF of a Gaussian pre-activation is
\begin{equation*}
    \phi_z(t) = \exp\!\big(\mathrm{i}mt - \tfrac{1}{2}s^2t^2\big),
\end{equation*}
where $m$ and $s^2$ are the mean and variance of the pre-activation, and where we have dropped the unit index $u$ for the sake of simplicity.
Completing the square in the exponent gives
\begin{equation}
    \phi_z(t) = e^{-v_0^2}\, e^{v(t)^2},
    \qquad v(t) = \frac{m + \mathrm{i}s^2t}{\sqrt{2}\,s},
    \qquad v_0 = v(0) = \frac{m}{\sqrt{2}\,s}.
    \label{eq:phi_z_square}
\end{equation}
Written in this form, $\phi_z(t)$ is a Gaussian, which has a known Hilbert transform
\begin{equation}
    \hilbert\big[e^{-t^2}\big](t) = \frac{2}{\sqrt{\pi}}\, D_+(t),
    \label{eq:hilert_gaussian_profile}
\end{equation}
where
\begin{equation*}
    D_+(t) = e^{-t^2} \int_0^t e^{y^2}\, dy
\end{equation*}
is the Dawson function.

The characteristic function in Eq.~\eqref{eq:phi_z_square} is shifted and rescaled from $e^{-t^2}$.
More precisely, we have that $\exp\bigl[v(t)^2\bigr] = \exp\bigl[-(-\mathrm{i}v(t))^2\bigr]$, where $-\mathrm{i}v(t) = st/\sqrt{2} - \mathrm{i}v_0$, which shows that $t$ is rescaled by a factor of $s/\sqrt{2}$ and shifted by $-\mathrm{i}v_0$.
These operations commute with the Hilbert transform, and the transform of $\phi_z(t)$ is therefore
\begin{equation*}
    \hilbert[\phi_z](t) = \frac{2}{\sqrt{\pi}}\, e^{-v_0^2}\, D_+\bigl(-\mathrm{i}v(t)\bigr),
\end{equation*}
where the leading term, $e^{-v_0^2}$, is the constant in Eq.~\eqref{eq:phi_z_square}, which passes through the transform by linearity.
Next, we use the identity 
\begin{equation*}
    D_+(-\mathrm{i}v) = -\mathrm{i}\frac{\sqrt{\pi}}{2}\, e^{v^2} \operatorname{erf}(v)
\end{equation*}
to obtain
\begin{equation}
    \hilbert[\phi_z](t) = -\mathrm{i}\, e^{v(t)^2 - v_0^2} \operatorname{erf}\big(v(t)\big) = -\mathrm{i}\, \phi_z(t) \operatorname{erf}\big(v(t)\big),
    \label{eq:hilbert_z_cf}
\end{equation}
which reduces to $\hilbert[\phi_z](0) = -\mathrm{i}\operatorname{erf}(v_0)$ at $t = 0$, since $\phi_z(0) = 1$.
Substituting both expressions into Eq.~\eqref{eq:relu_cf_app} and grouping like terms then yields
\begin{equation*}
    \phi_{\mathrm{ReLU}(z)}(t) = \frac{1}{2}\Big[ 1 - \operatorname{erf}(v_0) \Big] + \frac{1}{2}\,\phi_z(t)\Big[ 1 + \operatorname{erf}\bigl(v(t)\bigr) \Big].
\end{equation*}
Finally, we obtain  Eq.~\eqref{eq:relu_cf_gaussian} of the main text by using $\operatorname{erfc}(x) = 1 - \operatorname{erf}(x)$, and the fact that  $\operatorname{erf}$ is an odd function, allowing us to simplify the two brackets as:
\begin{equation*}
    \phi_{\mathrm{ReLU}(z)}(t) = \frac{1}{2}\operatorname{erfc}(v_0) + \frac{1}{2}\,\phi_z(t)\operatorname{erfc}\bigl(-v(t)\bigr).
\end{equation*}

This expression can be numerically unstable for large $|t|$, because $\phi_z(t)$ and $\operatorname{erfc}(-v(t))$ can individually underflow and overflow, even as their product remains finite.
Instead, our implementation uses the Faddeeva function
\begin{equation*}
    w(\zeta) = e^{-\zeta^2}\operatorname{erfc}(-\mathrm{i}\zeta),
\end{equation*}
to compute the product directly.
We find
\begin{equation*}
    \phi_{\mathrm{ReLU}(z)}(t) = \frac{1}{2}\operatorname{erfc}(v_0) + \frac{1}{2}\, e^{-v_0^2}\, w\big(-\mathrm{i}v(t)\big),
\end{equation*}
which can equivalently be written as 
\begin{equation*}
    \phi_{\mathrm{ReLU}(z)}(t) = \frac{1}{2}\operatorname{erfc}(v_0) + \phi_z(t) - \frac{1}{2}\, e^{-v_0^2}\, w\big(\mathrm{i}v(t)\big),
\end{equation*}
using the reflection identity $w(-\zeta) = 2e^{-\zeta^2} - w(\zeta)$.
In practice, we use the first form when $m \le 0$ and the second when $m > 0$, so that the argument of $w$ always falls in the upper half-plane, where $w$ is bounded.

\section{Derivation of the reconstruction costs}
\label{app:ell}

In this appendix, we derive the reconstruction costs shown in Table~\ref{tab:ell}.
To simplify the notation, we write $y$ for $y_n$, $\phi_\eta$ for $\phi_{\eta \mid x_n}$, and $\mathbb{E}_q$ for the expectation under $q(\eta \mid x_n)$ throughout the calculation.
The calculations below use the following three expectations, which are all consequences of Eq.~\eqref{eq:cf_expectation} in the main text:
\begin{equation}
    \mathbb{E}_q[\eta] = -\mathrm{i}\,\phi_\eta'(0), \qquad
    \mathbb{E}_q[\eta^2] = -\phi_\eta''(0), \qquad
    \mathbb{E}_q[e^{a\eta}] = \phi_\eta(-\mathrm{i}a).
    \label{eq:cf_moments}
\end{equation}
In all cases, the idea is to first compute the log-likelihood of the data for random outputs $\eta$, and to then take an expectation over $q$, substituting terms using Eq.~\eqref{eq:cf_moments}.

\subsection{Exponential-polynomial likelihoods}
For exponential-polynomial likelihoods conforming to Eq.~\eqref{eq:exp_sum_ll}, the reconstruction cost can be computed exactly using the characteristic function and its derivatives, typically with only a few terms.

\paragraph{Gaussian.}
For a Gaussian likelihood with variance $\sigma^2$, the log-likelihood is $\log p(y \mid \eta) = -\tfrac{1}{2}\log(2\pi\sigma^2) - (y-\eta)^2 / (2\sigma^2)$.
Expanding the square and applying Eq.~\eqref{eq:cf_moments} gives
\begin{equation}
    \mathbb{E}_q[\log p(y \mid \eta)] = -\frac{1}{2}\log(2\pi\sigma^2) - \frac{y^2 + 2\mathrm{i}\,y\,\phi_\eta'(0) - \phi_\eta''(0)}{2\sigma^2}.
\end{equation}

\paragraph{Poisson.}
For a Poisson likelihood with a log link, the rate is $e^{\eta}$ and the log-likelihood is $\log p(y \mid \eta) = y\eta - e^{\eta} - \log y!$, so we directly have
\begin{equation}
    \mathbb{E}_q[\log p(y \mid \eta)] = -\mathrm{i}\,y\,\phi_\eta'(0) - \phi_\eta(-\mathrm{i}) - \log y!.
\end{equation}

\paragraph{Gamma.}
For a Gamma likelihood with shape $\nu$ and mean $e^{\eta}$, the log-likelihood is $\log p(y \mid \eta) = \nu \log \nu - \log \Gamma(\nu) + (\nu - 1)\log y - \nu\eta - \nu y e^{-\eta}$, so we directly have
\begin{equation}
    \mathbb{E}_q[\log p(y \mid \eta)] = \nu \log \nu - \log \Gamma(\nu) + (\nu-1)\log y + \mathrm{i}\,\nu\,\phi_\eta'(0) - \nu\,y\,\phi_\eta(\mathrm{i}).
\end{equation}

\paragraph{Weibull.}
For a Weibull likelihood with shape $\kappa$ and scale $e^{\eta}$, the log-likelihood is $\log p(y \mid \eta) = \log\kappa + (\kappa-1)\log y - \kappa\eta - y^{\kappa} e^{-\kappa\eta}$, so we directly have
\begin{equation}
    \mathbb{E}_q[\log p(y \mid \eta)] = \log\kappa + (\kappa-1)\log y + \mathrm{i}\,\kappa\,\phi_\eta'(0) - y^{\kappa}\,\phi_\eta(\mathrm{i}\kappa).
\end{equation}

\paragraph{General exponential families.}
The same computation can be applied to many exponential families.
In the parametrization in terms of sufficient statistics $T_j(y)$, natural parameters $\theta_j(\eta)$, and log-partition function $A(\eta)$, the expected log-likelihood is
\begin{equation}
    \mathbb{E}_q[\log p(y \mid \eta)] = \log h(y) - \mathbb{E}_q[A(\eta)]  + \sum_j T_j(y)\,\mathbb{E}_q[\theta_j(\eta)].
\end{equation}
When $A$ and all the natural parameters $\theta_j$ are sums of terms $\eta^{r} e^{a\eta}$, the expectation has a closed form.

\subsection{Categorical likelihood}

In classification problems, the output label is drawn from a categorical distribution whose probabilities are the softmax of the network outputs.
Hence, the likelihood can be calculated as $\log p(y \mid \boldsymbol{\eta}) = \eta_y - \log \sum_{k'} e^{\eta_{k'}}$.
This likelihood is, unfortunately, not an exponential-polynomial, because the logarithm of a sum of exponentials cannot be rewritten as a finite sum of terms of the form $\eta^{r} e^{a\eta}$.

Instead, following \citet{harrison_variational_2024}, we obtain a lower bound on the expected log-likelihood, using Jensen's inequality:
\begin{equation}
    \mathbb{E}_q[\log p(y \mid \boldsymbol{\eta})] \;\ge\; \mathbb{E}_q[\eta_y] - \log \sum_{k'} \mathbb{E}_q\!\left[e^{\eta_{k'}}\right].
\end{equation}
The bound moves the expectation inside the logarithm, where the terms $\mathbb{E}_q[e^{\eta_{k'}}]$ involve a single output and are given exactly by Eq.~\eqref{eq:cf_moments}.
Applying Eq.~\eqref{eq:cf_moments} to the lower bound then gives
\begin{equation}
    \mathbb{E}_q[\log p(y \mid \boldsymbol{\eta})] \;\ge\; -\mathrm{i}\,\phi_{\eta_y}'(0) - \log \sum_{k'} \phi_{\eta_{k'}}(-\mathrm{i}).
\end{equation}

\subsection{General likelihoods}

When the reconstruction cost has no closed form, we must invert the CF on a grid with Eq.~\eqref{eq:cf_inversion} and evaluate the integral in Eq.~\eqref{eq:likelihood_cost_integral} by quadrature:
\begin{equation}
    \mathbb{E}_q[\log p(y \mid \eta)] \approx \Delta\eta \sum_{j=1}^{G} \log p(y \mid \eta_j)\, q(\eta_j \mid x_n).
\end{equation}
Gradients can be taken through this deterministic transformation.
We use an adaptive grid of $G$ points spanning $\pm 10$ standard deviations around the mean of $\eta$, which can lead to mild numerical artifacts for heavy-tailed output distributions.

\end{document}